\documentclass[12pt,letterpaper]{article}
\usepackage[utf8]{inputenc}
\usepackage[margin=1in]{geometry}
\usepackage{graphicx}
\usepackage{amsmath}
\usepackage{amssymb}
\usepackage{booktabs}
\usepackage{xcolor}
\usepackage{pdfpages}   % in preamble

\usepackage{hyperref}
\usepackage{xurl}
\usepackage{natbib}
\usepackage{setspace}
\usepackage{float}
\usepackage[normalem]{ulem}
\newcommand{\new}[1]{#1}
\title{\textbf{AGAPI-Agents: An Open-Access Agentic AI Platform for Accelerated Materials Design on AtomGPT.org}}

\author{
Jaehyung Lee$^{1}$, Justin Ely$^{2}$, Kent Zhang$^{2}$, Akshaya Ajith$^{1}$, \\
Charles Rhys Campbell$^{1,3}$, Kamal Choudhary$^{1,2,*}$ \\
\\
\scriptsize $^{1}$Department of Materials Science and Engineering, Johns Hopkins University, Baltimore, MD 21218, USA \\
\scriptsize $^{2}$Department of Electrical and Computer Engineering, Johns Hopkins University, Baltimore, MD 21218, USA \\
\scriptsize $^{3}$Department of Physics and Astronomy, West Virginia University, Morgantown, WV 26506, USA \\
\footnotesize $^{*}$Corresponding author: kchoudh2@jhu.edu
}
\date{}

\begin{document}

\maketitle
\begin{abstract}
Agentic AI systems increasingly connect large language models (LLMs) to external scientific tools, yet whether \new{and when} tool access improves prediction accuracy remains uncharacterized. \new{We present AGAPI (AtomGPT.org API), an open-access platform integrating eight open-source LLMs with 18 REST endpoints (28 agent tools, 50 web apps) spanning materials databases, force fields, tight-binding band structures, X-ray diffraction, and protein structure. A three-evaluation residual decomposition on JARVIS-Leaderboard electronic-structure test sets separates agent pipeline fidelity from inherited density functional theory (DFT) functional bias. For bulk modulus and bandgap the agent reproduces JARVIS-DFT entries to numerical precision, so the experimental-reference degradation is functional bias, not agentic malfunction. On memorization-resistant test sets (57 defective supercells, 60 hypothetical compositions), tool-augmented mean absolute error (MAE) is below $0.005$~eV versus $1.25$ to $1.86$~eV tool-free, confirming tools are indispensable where parametric knowledge is unavailable.} We further demonstrate autonomous multi-step workflows including 10-operation defect-engineering pipelines. AGAPI is available at \url{https://github.com/atomgptlab/agapi}.
\end{abstract}

\section*{TOC Graphic}
\begin{center}
\includegraphics[width=2in,keepaspectratio]{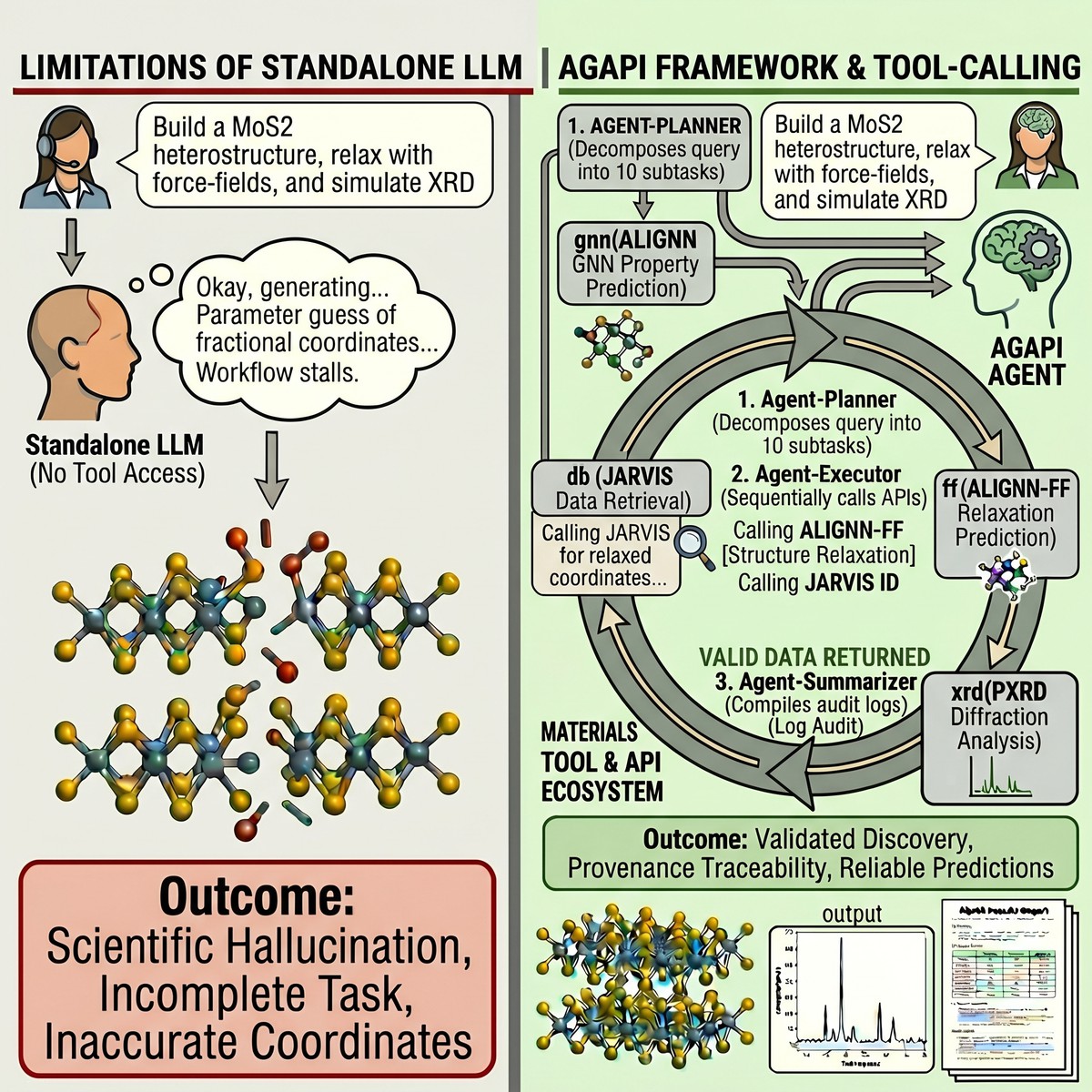}
\end{center}

The accelerating pace of scientific discovery increasingly demands the integration of heterogeneous computational tools, expansive databases, and sophisticated machine learning models into coherent workflows.\cite{wang2023scientific,choudhary2022recent,wilkinson2016fair,zitnik2019machine} Large language models (LLMs) have emerged as promising orchestrators for such workflows, demonstrating capabilities in natural language understanding, multi-step reasoning, and code generation.\cite{kasneci2023chatgpt,naveed2025comprehensive,achiam2023gpt} In materials science, LLMs show potential for tasks ranging from literature synthesis and experimental design to property prediction and inverse materials discovery.\cite{jablonka202314,jablonka2024leveraging,choudhary2023chemnlp,choudhary2025chatgpt,choudhary2025diffractgpt,choudhary2024atomgpt,choudhary2025microscopygpt,antunes2024crystal}

Current approaches to incorporating materials science knowledge in LLMs fall into three categories: training from scratch on scientific corpora,\cite{taylor2022galactica,flam2023language} fine-tuning pre-trained models on domain-specific datasets,\cite{choudhary2024atomgpt,rubungo2023llm,gruver2024fine,antunes2024crystal,choudhary2025diffractgpt,choudhary2025microscopygpt} and developing agentic frameworks that augment LLMs with external tools.\cite{choudhary2025chatgpt,bran2023chemcrow,boiko2023autonomous} Training from scratch requires computational resources beyond most research groups. Fine-tuning has shown success but requires curated datasets for each application domain and lacks flexibility across material classes. Agentic AI, defined here as single AI agents equipped with multiple external tools that can autonomously plan and execute multi-step tasks, represents a complementary paradigm that leverages pre-trained LLMs as reasoning engines connected to databases, simulation codes, and machine-learning models through orchestrated workflows.\cite{wei2025ai,qin2023toolllm,schick2023toolformer,padgham2005developing}

Notable agentic frameworks include Coscientist for autonomous chemical experimentation,\cite{boiko2023autonomous} ChemCrow for multi-tool chemistry workflows,\cite{bran2023chemcrow} AtomAgents for materials simulations,\cite{ghafarollahi2024atomagents} AURA for NanoHub integration,\cite{gastelum2025autonomous} LLamp for Materials Project integration,\cite{chiang2024llamp} SciToolAgent for knowledge-graph-guided tool use,\cite{ding2025scitoolagent} and ChatGPT Material Explorer for JARVIS integration in ChatGPT.\cite{choudhary2025chatgpt} However, a fundamental question remains largely unaddressed: \textit{when does connecting an LLM to external tools actually improve the accuracy of scientific predictions, and when does it introduce errors?} Tool augmentation is implicitly assumed to be beneficial, yet several factors complicate this assumption. First, databases may contain systematic errors. For example, electronic bandgaps computed with standard DFT functionals are known to underestimate experimental values by 30--50\%.\cite{choudhary2020joint} Second, for well-characterized materials extensively covered in textbooks and literature, an LLM's parametric knowledge may already be highly accurate, and tool access can introduce noise through polymorph mismatches or retrieval of inferior computational values. Third, error propagation in multi-step tool chains can compound inaccuracies across sequential operations. Moreover, most existing agentic frameworks rely on commercial LLMs (e.g., GPT-4, Claude), limiting reproducibility due to non-deterministic behavior across unannounced API version updates.\cite{liang2022holistic}

Here we address these gaps using AGAPI (AtomGPT.org API),\cite{choudhary2025chatgpt} an open-access agentic AI platform that integrates eight open-source LLMs with \new{18 REST API endpoints (wrapped as 28 LLM-callable agent tools)} spanning \new{the Joint Automated Repository for Various Integrated Simulations DFT database (JARVIS-DFT\cite{choudhary2020joint,wines2023recent,choudhary2025jarvis,choudhary2021high,garrity2021database})}, \new{the Atomistic Line Graph Neural Network (ALIGNN\cite{choudhary2021atomistic,gurunathan2023rapid}) for property prediction}, machine-learning force fields (ALIGNN-FF,\cite{choudhary2023unified} CHIPS-FF\cite{wines2025chips}), tight-binding band structure calculations (SlaKoNet\cite{choudhary2025slakonet}), X-ray diffraction \new{(XRD)} analysis (DiffractGPT\cite{choudhary2025diffractgpt}), and protein structure prediction (ESMFold\cite{lin2023evolutionary}). AGAPI employs a single-agent architecture with \new{a Planner-Executor-Summarizer reasoning loop} built on the OpenAI Agents SDK\cite{openai_agents_python}. \new{The agent dispatches to multiple tools rather than coordinating task-specific sub-agents.} All LLM inference is performed on self-hosted servers using vLLM\cite{kwon2023efficient} and Ollama,\cite{ollama} ensuring complete reproducibility through version pinning and deterministic sampling. \new{We compare tool-augmented versus tool-free materials property predictions, scoring each against both DFT and experimental references. This separates two sources of error: bias in the underlying DFT functional, and error introduced by the agent pipeline itself. A residual decomposition separates database-vs-experiment error from agent-vs-database error, showing that the observed deterioration on well-characterized properties is not a failure of tool augmentation as a paradigm, but a statement about database-functional mismatch for specific properties. To our knowledge, AGAPI provides the first quantitative side-by-side comparison of tool-augmented vs.\ tool-free predictions for crystalline materials spanning electronic, elastic, dielectric, optical, and superconducting properties, evaluated against both DFT and experiment. Table~\ref{tab:platforms} positions this coverage relative to existing agentic frameworks for materials and chemistry.}

\begin{table}[H]
\centering
\caption{\new{Comparison with prior agentic platforms for materials science and chemistry across seven axes: (1) open-source LLM support, (2) reproducibility via version pinning, (3) number of integrated materials-science tools, (4) coverage of hard and soft matter, (5) a public REST API, (6) quantitative tool-augmented vs.\ tool-free evaluation on property-prediction or simulation tasks with measurable reference standards, and (7) evaluation against experimental data. ``API-dep.'' indicates that reproducibility depends on a commercial model-provider API. ChatGPT-MX = ChatGPT Material Explorer.\cite{choudhary2025chatgpt} Platform references: ChemCrow,\cite{bran2023chemcrow} Coscientist,\cite{boiko2023autonomous} AtomAgents,\cite{ghafarollahi2024atomagents} AURA,\cite{gastelum2025autonomous} LLamp,\cite{chiang2024llamp} SciToolAgent.\cite{ding2025scitoolagent} SciToolAgent's tool count reflects general-purpose knowledge-graph routing rather than curated materials-science APIs.}}
\label{tab:platforms}
\vspace{0.8em}
\scriptsize
\resizebox{\textwidth}{!}{%
\begin{tabular}{lcccccccc}
\toprule
\textbf{Axis} & \textbf{AGAPI} & \textbf{ChemCrow} & \textbf{Coscientist} & \textbf{AtomAgents} & \textbf{AURA} & \textbf{LLamp} & \textbf{SciToolAgent} & \textbf{ChatGPT-MX} \\
\midrule
Open-source LLM & \checkmark & -- & -- & -- & -- & -- & -- & -- \\
Version-pinned reprod. & \checkmark & API-dep. & API-dep. & API-dep. & API-dep. & API-dep. & API-dep. & API-dep. \\
\# MS agent tools integrated & 28 & 18 & $\sim$5 & $\sim$8 & 340+ (gen.) & $\sim$5 & 500+ (gen.) & $\sim$10 \\
Hard $+$ soft matter & \checkmark & Soft only & Soft only & Hard only & General & Hard only & General & Hard only \\
Public REST API & \checkmark & -- & -- & -- & Partial & -- & Partial & Partial \\
Tool-aug.\ vs.\ tool-free eval. & \checkmark & \checkmark & Limited & -- & -- & \checkmark & -- & -- \\
Eval.\ against experiment & \checkmark & Limited & Limited & -- & -- & -- & -- & Limited \\
\bottomrule
\end{tabular}%
}
\end{table}

\begin{figure}[H]
\centering
\includegraphics[width=\textwidth]{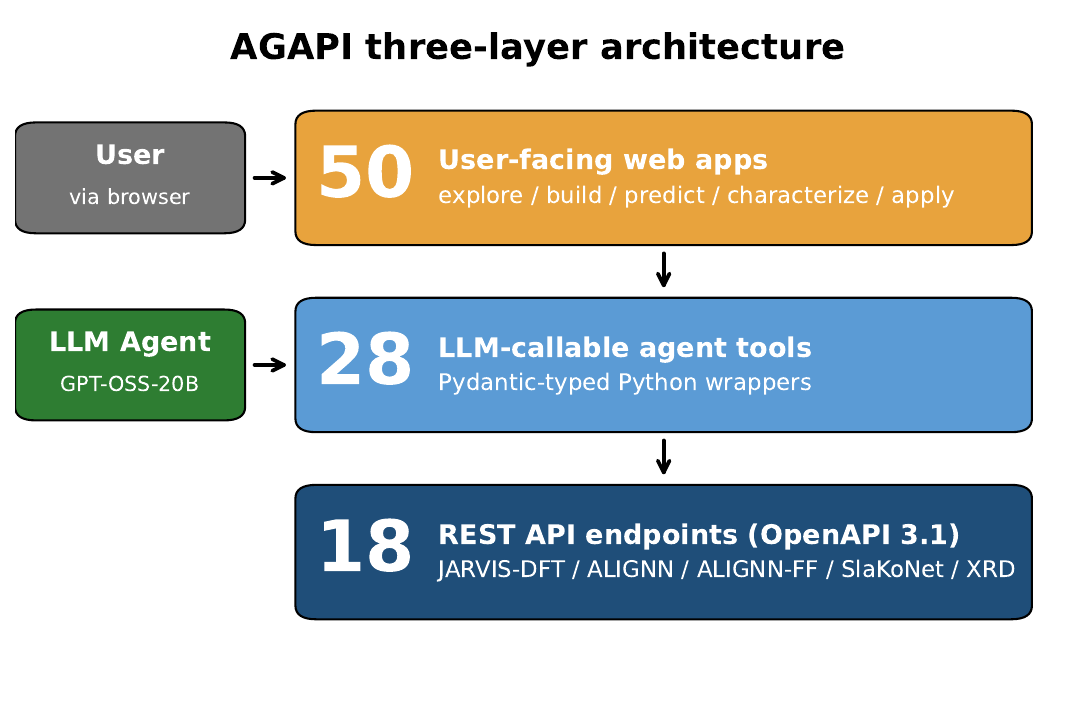}
\caption{\new{AGAPI three-layer architecture. The 50 user-facing web apps on AtomGPT.org (top) and the 28 LLM-callable Pydantic-typed agent tools (middle) both dispatch to the same 18 OpenAPI 3.1 REST endpoints (bottom), which wrap databases, ML models, and simulation tools (JARVIS-DFT, ALIGNN, ALIGNN-FF, SlaKoNet, XRD). Users enter via the web-app layer and the LLM agent (default GPT-OSS-20B) enters via the tool layer.}}
\label{fig:arch}
\end{figure}

\textbf{Platform Architecture.} \new{AGAPI is organized as three stacked layers (Figure~\ref{fig:arch}): 50 user-facing web apps for browser access, 28 LLM-callable agent tools (Pydantic-typed Python wrappers), and 18 OpenAPI 3.1 REST endpoints that the upper layers dispatch to. The User enters via the web-app layer and the LLM agent enters via the tool layer.} This separation enables independent improvement of both components: as more capable open-source LLMs emerge, AGAPI can integrate them without modifying the tool layer, and conversely, new databases, simulation methods, or ML models can be added as API endpoints without retraining or reconfiguring the LLM.

\new{For clarity in what follows we distinguish three layers whose sizes are often conflated in descriptions of agentic platforms. (i) \textit{REST API endpoints}: the 18 OpenAPI-3.1 GET/POST routes that the agent's tool wrappers dispatch to. (ii) \textit{Agent tools}: the 28 Python-side tool wrappers, each defined by a Pydantic schema that the LLM selects from during planning. Some agent tools wrap multiple endpoints or add pre/post-processing, which is why the tool count exceeds the endpoint count. (iii) \textit{Web apps}: the 50 user-facing interactive applications hosted on AtomGPT.org that consume the same REST layer as the agent but are not themselves LLM-callable. Unless noted otherwise, ``tools'' in this manuscript refers to agent tools (the layer the LLM sees), and all API counts refer to REST endpoints.}

\new{Within the agent-tool layer, a single agent's reasoning loop runs in three sequential phases (Planner, Executor, Summarizer). User queries enter through natural-language interfaces (web chatbot, Python API, or voice input). The Planner analyzes the query to identify required tools, data sources, and workflows, decomposing complex requests into task sequences with explicit dependencies. The Executor dispatches API calls asynchronously, handles rate limiting, and implements retry logic for transient failures. The Summarizer aggregates results, validates physical consistency, and generates formatted outputs including tables, plots, and atomic-structure visualizations.}

\new{The 18 REST endpoints (exposed to the LLM through 28 agent-tool wrappers)} are organized into functional modules. \textit{Database Query Endpoints} provide structured access to JARVIS-DFT,\cite{choudhary2020joint} Materials Project,\cite{jain2013commentary} the Automatic FLOW for Materials Discovery database (AFLOW),\cite{curtarolo2012aflow} and the Open Quantum Materials Database (OQMD)\cite{kirklin2015open} via the Open Databases Integration for Materials Design (OPTIMADE) API,\cite{andersen2021optimade} supporting conjunctive queries across numerous properties including composition, bandgap ranges, formation energy, and elastic constants. \textit{Property Prediction Endpoints} leverage pre-trained ALIGNN\cite{choudhary2021atomistic} graph neural network models for rapid estimation of formation energy, bandgaps from both the modified Becke-Johnson (MBJ) and OptB88vdW functionals, elastic constants, dielectric properties, and superconducting critical temperature from crystal structures in POSCAR format. \textit{Force Field Endpoints} implement ALIGNN-FF\cite{choudhary2023unified} and CHIPS-FF\cite{wines2025chips} for near-DFT accuracy structure optimization with automatic convergence detection. \textit{Structure Generation Endpoints} support building supercells, point defects (vacancies, substitutions), and heterostructure interfaces using the Zur algorithm for coincidence site lattice matching. \textit{Characterization Endpoints} simulate powder XRD patterns using specified radiation sources (Cu K$\alpha$) and compute electronic band structures via the SlaKoNet\cite{choudhary2025slakonet} tight-binding framework\new{, benchmarked alongside conventional TB parametrizations in CHIPS-TB.\cite{park2026chipstb}} \textit{Protein Structure Endpoints} provide access to ESMFold,\cite{lin2023evolutionary} the Protein Data Bank (PDB),\cite{altschul1990protein} and AlphaFold\cite{varadi2022alphafold} databases. All endpoints implement authentication via JWT tokens and rate limiting via a token bucket algorithm. \new{Tool integration is lightweight: 30--50 lines of Python for the API endpoint, and 10--20 lines for registry metadata.} The LLM discovers new tools automatically from schema descriptions without requiring any changes to the agent code.

\textbf{Open-Source LLM Evaluation.} A critical design decision in AGAPI was selecting open-source LLMs that balance reasoning capability, inference speed, and accessibility. We evaluated eight models on token generation speed, a key metric for interactive scientific workflows (Figure~S5a,b). Models span a range of sizes and architectures: Llama-3.2-90B-Vision\new{\cite{grattafiori2024llama3}} (multimodal, 90B parameters), DeepSeek-V3\cite{liu2024deepseek} (mixture-of-experts), Qwen3-Next-80B\new{\cite{yang2025qwen3}} (80B), Gemma-3-27B\new{\cite{team2025gemma3}} (27B), Kimi-K2\new{\cite{team2025kimi2}} (reasoning-focused), GPT-OSS-20B and GPT-OSS-120B\cite{gpt_oss_20b} (OpenAI open-source releases), and Phi-4\cite{abdin2024phi} (small but capable). All benchmarks use text-only input. Multimodal capabilities of vision-augmented models are relevant for planned extensions such as microscopy image analysis but are not exercised in the current evaluation.

Using Llama-3.2-90B-Vision as the baseline (36.1 tokens/s), we observed substantial speedups with GPT-OSS-20B (141.7 tokens/s, 3.93$\times$) and GPT-OSS-120B (122.3 tokens/s, 3.39$\times$). Mid-tier models achieved moderate acceleration: Qwen3-Next-80B (95.8 tokens/s, 2.66$\times$) and Kimi-K2 (53.3 tokens/s, 1.48$\times$). Vision-augmented models exhibited higher variance, reflecting computational overhead from vision encoder modules that remain active even during text-only inference. Load testing simulated 1{,}000 concurrent users with staggered request patterns, yielding a mean response time of 16.6~s at peak load (Figure~S5b), with planned improvements targeting sub-2-second responses through horizontal scaling.

Given that GPT-OSS-20B achieves the highest throughput while demonstrating strong performance on established benchmarks, namely the American Invitational Mathematics Examination (AIME), Graduate-Level Google-Proof Q\&A (GPQA), and Massive Multitask Language Understanding (MMLU),\cite{gpt_oss_20b} we selected it as the default model for the agentic infrastructure. Users may choose other available models or integrate commercial APIs when needed. \new{Users interact with the selected LLM through two AtomGPT.org entry points: 50 web apps organized by research workflow stage (explore, build, predict, characterize, apply, validate) (Figure~S5c), and a chatbot interface that performs tool-calling grounded in actual database queries (Figure~S5d).}

\textbf{Tool-Augmented vs.\ Tool-Free Predictions.} \new{We now turn to the central question motivating this work: does connecting an LLM to physics-based tools improve the accuracy of materials property predictions? We benchmark GPT-OSS-20B at temperature $= 0$ (each query independent, no conversation history) on five JARVIS-Leaderboard\cite{choudhary2024jarvis} ES test sets: bandgap ($N = 55$), bulk modulus ($N = 45$), superconducting $T_c$ ($N = 38$), solar spectroscopic limited maximum efficiency (SLME, $N = 28$), and dielectric constant $\varepsilon_x$ ($N = 32$).}

\new{Against experimental references, tool augmentation improves two of these properties: bulk modulus by 27\% (MAE $7.88 \rightarrow 5.73$~GPa, $R^2$: $0.984 \rightarrow 0.994$, Figure~S6b) and dielectric constant $\varepsilon_x$ by 46\% (MAE $2.88 \rightarrow 1.54$, $R^2$: $0.06 \rightarrow 0.78$, Figure~S6e), confirming the value of tool retrieval where the underlying DFT reference is accurate or the LLM has no strong parametric prior. For the remaining three properties, the tool-augmented MAE against experiment is larger than the tool-free MAE (bandgap by 40\%, $T_c$ more than fivefold, SLME by 63\%, Figure~S6a,c,d).}

\new{The central question is whether the larger MAE on these three properties comes from the agent itself or from the DFT values it retrieves. To answer it, we structure the analysis as three complementary evaluations (A, B, C) on the subset of materials where $V_{\text{agent}}$ could be reliably extracted from the agent's natural-language response. Evals A and C cover bandgap, bulk modulus, and $T_c$, while Eval B is specific to bandgap. Per-property $n$ values are reported with each evaluation below. For each material and property we compare three values of the same quantity, distinguished only by their source: $V_{\text{exp}}$ (the experimental reference from the JARVIS-Leaderboard ES test set), $V_{\text{db}}$ (the DFT-computed value retrieved directly from the JARVIS-DFT REST endpoint), and $V_{\text{agent}}$ (the value reported by the AGAPI agent). The three pairwise differences are the agent-vs-database residual ($V_{\text{agent}} - V_{\text{db}}$, agentic pipeline fidelity), the database-vs-experiment residual ($V_{\text{db}} - V_{\text{exp}}$, inherited DFT functional bias), and their sum, the total tool-augmented error against experiment ($V_{\text{agent}} - V_{\text{exp}}$).}

\begin{figure}[H]
\centering
\includegraphics[width=\textwidth]{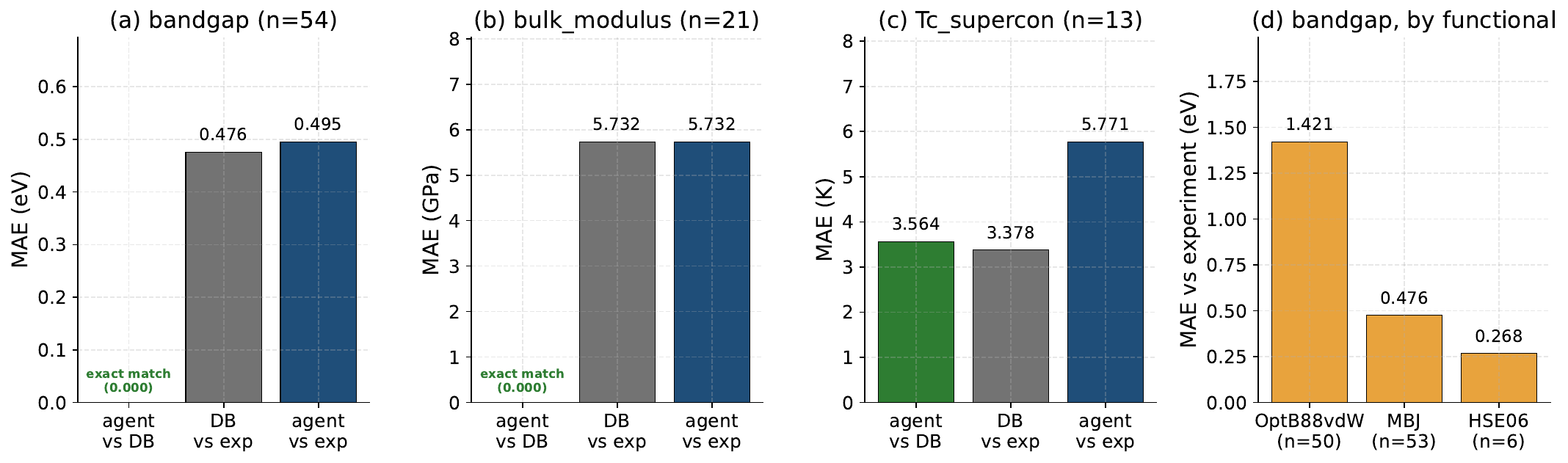}
\caption{\new{Three-evaluation analysis of tool-augmented error on the JARVIS-Leaderboard ES test sets. (a--c) Per-property MAEs , agent-vs-DB (green, Evaluation~A pipeline fidelity), DB-vs-exp (gray, inherited DFT functional bias), and agent-vs-exp (blue, total tool-augmented error against experiment) , for bandgap ($n = 54$), bulk modulus ($n = 21$), and $T_c$ ($n = 13$). (d) Evaluation~B: bandgap MAE against experiment stratified by DFT functional , OptB88vdW ($n = 50$), MBJ ($n = 53$), and HSE06 ($n = 6$). Evaluation~C is the joint reading of panels (a--c).}}
\label{fig:decomp}
\end{figure}

\new{\textit{Evaluation~A: Pipeline fidelity against DFT reference values.} Scoring the tool-augmented predictions against the same DFT database entries the agent retrieves isolates agentic behavior from functional accuracy. For bulk modulus ($n = 21$ from the JARVIS-Leaderboard \texttt{dft\_3d\_bulk\_modulus} test set) and bandgap ($n = 54$ from \texttt{dft\_3d\_bandgap}, with $V_{\text{db}}$ successfully retrieved for 53 of 54, the exception being Cu$_2$O for which JARVIS-DFT has no MBJ bandgap entry), the agent's reported values agree with the retrieved JARVIS-DFT entries to numerical precision on every material where $V_{\text{db}}$ was available: $\mathrm{MAE}(V_{\text{agent}}, V_{\text{db}}) = 0.0$~GPa ($n = 21$) and $0.0$~eV ($n = 53$) respectively (Figure~\ref{fig:decomp}a--b). For these two properties the agentic pipeline (tool selection, argument passing, result handling) reproduces the underlying tool output without modification, so any error against experiment originates in the database itself. Results for bulk modulus and $T_c$ are reported against the corresponding DFT-derived reference in JARVIS-DFT:\cite{choudhary2020joint} the Voigt-averaged bulk modulus from the OptB88vdW elastic tensor, and the $T_c$ scalar from DFT electron--phonon coupling (McMillan--Allen--Dynes formula with Coulomb pseudopotential $\mu^* = 0.1$) for conventional, phonon-mediated superconductors. For superconducting $T_c$ ($n = 13$ from \texttt{dft\_3d\_Tc\_supercon}) the agent-vs-database residual is nontrivial ($\mathrm{MAE} = 3.56$~K, Figure~\ref{fig:decomp}c), indicating that on this property the agent does not always forward the retrieved DFT electron--phonon value verbatim. Unlike bandgap and bulk modulus, the $T_c$ decomposition has both DFT-bias and agent contributions. This $T_c$ behavior is consistent with the agent operating as a hybrid system rather than a passive retrieval pipeline. Making this hybrid behavior consistent and explicit is the motivation for the adaptive-tool-selection scheme discussed below.}

\new{\textit{Evaluation~B: Experimental reference, stratified by functional.} For bandgap we report tool-augmented errors separately against three functionals: OptB88vdW ($\mathrm{MAE} = 1.42$~eV, $n = 50$), MBJ ($\mathrm{MAE} = 0.48$~eV, $n = 53$), and HSE06 ($\mathrm{MAE} = 0.27$~eV, $n = 6$) (Figure~\ref{fig:decomp}d). OptB88vdW, MBJ, and HSE06 values are pulled directly from each material's JARVIS-DFT entry. The monotonic drop directly demonstrates that what is loosely called ``tool error'' on bandgap is, in large part, functional-choice error inherited by the agent: switching the retrieved reference from OptB88vdW to HSE06 reduces the experimental-reference error by a factor of $\sim$5 without changing the agent.}

\new{\textit{Evaluation~C: Residual decomposition.} The tool-augmented error against experiment has two contributing sources: a database-vs-experiment residual (DFT functional bias inherited from JARVIS-DFT) and an agent-vs-database residual (the agent's own contribution). For bulk modulus and bandgap, Evaluation~A showed the agent forwards the database value exactly, so the total error against experiment, $5.73$~GPa and $0.495$~eV respectively, is entirely DFT functional bias (Figure~\ref{fig:decomp}a--b). For $T_c$, both sources contribute comparably: $3.38$~K from DFT, $3.56$~K from the agent, $5.77$~K total (Figure~\ref{fig:decomp}c).}

\textbf{Failure-Mode Analysis.} \new{The residual decomposition of Evaluation~C reframes the three failure mechanisms below as \textit{sources of the database-vs-experiment residual} (DFT functional bias inherited by the agent), rather than as failures of tool augmentation. For $T_c$, which is the one property with a nonzero agent-vs-database residual, the agent-side contribution was noted in Evaluation~A, so the discussion here addresses only the DFT-bias side.}

\textit{Mechanism 1: Electron--phonon DFT for $T_c$.} JARVIS-DFT stores superconducting $T_c$ from DFT electron--phonon coupling calculations, which capture the correct physics but are sensitive to $k$-point sampling, phonon frequency accuracy, and the exchange-correlation functional. For well-studied superconductors documented in textbooks, review articles, and Wikipedia, the DFT value is less accurate than the literature value encoded in the LLM's parametric knowledge. For MgB$_2$ the database entry is 32.7~K versus the experimental $\sim$39~K. For Nb$_3$Sn the electron--phonon estimate deviates further from experiment ($\approx$18~K) than the LLM's parametric estimate. \new{Figure~\ref{fig:decomp}c shows that this database-vs-experiment offset contributes substantially to the tool-augmented error on $T_c$. On this property, the agent does not just forward the DFT value. The agent-vs-database residual is comparable in size to the database-vs-experiment residual (Evaluation~A above), meaning the agent sometimes replaces the retrieved DFT electron--phonon $T_c$ with a value from the LLM's parametric memory. For canonical superconductors that memorized value is often closer to experiment than the DFT estimate.}

\textit{Mechanism 2: Meta-GGA bandgap underestimation.} The agent retrieves MBJ meta-GGA bandgaps, which substantially improve on GGA but still systematically underestimate experimental bandgaps for many material classes.\cite{choudhary2020joint} \new{Figure~\ref{fig:decomp}a confirms that the residual error on bandgap is entirely database-vs-experiment (the agent-vs-database residual is zero on all 54 materials evaluated). This mechanism is addressable by replacing the retrieved reference with hybrid-functional or $GW$ entries rather than by changing the agent.}

\textit{Mechanism 3: Error amplification for derived properties.} \new{SLME is a derived quantity computed from the optical absorption spectrum and bandgap, so upstream DFT errors propagate multiplicatively into the tool-augmented prediction. For the dielectric constant $\varepsilon_x$ (also derived, from DFPT) the comparison goes the other way: tool augmentation reduces MAE against experiment by 46\% (Figure~S6e), because the LLM has no strong memorized prior for specific dielectric constants and the DFPT estimate is closer to experiment than the LLM's tool-free guess.} Detailed benchmarking of DFT accuracy for each of these properties against experimental data is provided in the respective JARVIS publications.\cite{choudhary2020joint,wines2023recent,choudhary2024jarvis}

\new{Taken together, these three mechanisms locate the source of the larger MAE in the DFT entries the agent retrieves, not in the act of using tools.} Its net benefit depends on three factors: (1) the accuracy of the DFT methodology for the target property, (2) whether the DFT-computed values are systematically biased relative to experiment, and (3) whether the material and its properties are well-documented in the LLM's training corpus. For well-known materials the LLM's parametric knowledge, distilled from the entire scientific literature during pre-training, can be more accurate than DFT-computed values in a specific database, a pattern consistent with recent work on tool-use evaluation in the broader LLM literature\cite{ning2024wtu,shen2024smartcal,ma2025advancing} showing that LLMs can outperform tool-augmented approaches on well-characterized problems.

\new{\textit{Quantitative support for the three mechanisms.} To move the failure-mode discussion beyond plausible interpretation we report two quantitative analyses on the 54-material JARVIS-Leaderboard ES bandgap test set (Figure~S7). These complement the functional comparison already reported as Evaluation~B above (Figure~\ref{fig:decomp}d), which directly demonstrates Mechanism~2 (meta-GGA bandgap underestimation). Both analyses partition the test set into subgroups and report, for each subgroup, the agent-vs-experiment MAE averaged over the materials in that subgroup.}

\new{\textit{(a) Material-class stratification} (Figure~S7a). We classified each test-set material by a composition-based heuristic into seven classes (oxide, halide, carbide, nitride, chalcogenide, 2D, and other). Per-class agent-vs-experiment MAE varies markedly: oxides 1.04~eV ($n = 9$; Cu$_2$O excluded as in Evaluation~A), halides 1.05~eV ($n = 6$), carbides 0.46~eV ($n = 1$), nitrides 0.38~eV ($n = 5$), chalcogenides 0.33~eV ($n = 14$), 2D materials 0.17~eV ($n = 6$), other 0.13~eV ($n = 12$). The wide-bandgap classes (oxides and halides) have the largest error, consistent with the well-known systematic underestimation of large gaps by semilocal and meta-GGA functionals.}

\new{\textit{(b) Data-availability stratification} (Figure~S7b). If the LLM has memorized property values from its pre-training data, then citation count should predict tool-free accuracy: high-citation materials provide a strong memorized prior (low tool-free MAE expected) while rarely-cited materials provide a weak one (high tool-free MAE expected). We queried Crossref for the publication mention count of each formula and binned materials by mentions: $[0, 10)$, $[10, 100)$, $[100, 1000)$, and $[1000, \infty)$. The 54-material set is heavily skewed: 50 materials fall in the $\geq 1000$ bin, 2 in the $[100, 1000)$ bin (MAE 0.50~eV), 0 in the $[10, 100)$ bin, and 2 in the $<10$ bin, so the per-bin MAE in the lower bins is underpowered. The directional finding (low-mention materials show MAE 0.70~eV vs 0.47~eV in the heavily-cited bin) is consistent with this parametric-memorization prediction.}

\new{\textbf{Memorization-Resistant Evaluation.} To test the complementary claim that tool augmentation is essential where parametric knowledge is unavailable, we constructed two purpose-built test sets designed to fall outside the LLM's training distribution. DEF-57 (defective supercells) comprises 57 defective configurations across five canonical hosts (Si, GaN, MgO, Al$_2$O$_3$, NaCl), with substitutional dopants, vacancies, and antisite swaps generated locally via jarvis-tools (the tool-augmented agent returned a numeric value for 57 of 57, of which 56 have a matching ALIGNN ground truth and contribute to the reported MAE). HYP-60 (hypothetical compositions) comprises 60 novel compositions generated by prototype substitution from six well-known crystal structures (Si, GaN, MgO, ZnS, BN, LiF), with all atoms of the target element replaced by chemically plausible substituents. For each structure we obtain a ground-truth bandgap from ALIGNN and compare the predictions of the AGAPI agent under two conditions: with the tool layer enabled (tool-augmented) and with it disabled (tool-free). The tool-augmented MAE is on the diagonal by construction (the agent calls ALIGNN under the hood, which is also the ground-truth source). The independent evidence comes from the tool-free side. With no tool access, the LLM must predict bandgaps from memory alone, on defective and hypothetical structures specifically constructed to fall outside its training distribution.}

\begin{figure}[H]
\centering
\includegraphics[width=\textwidth]{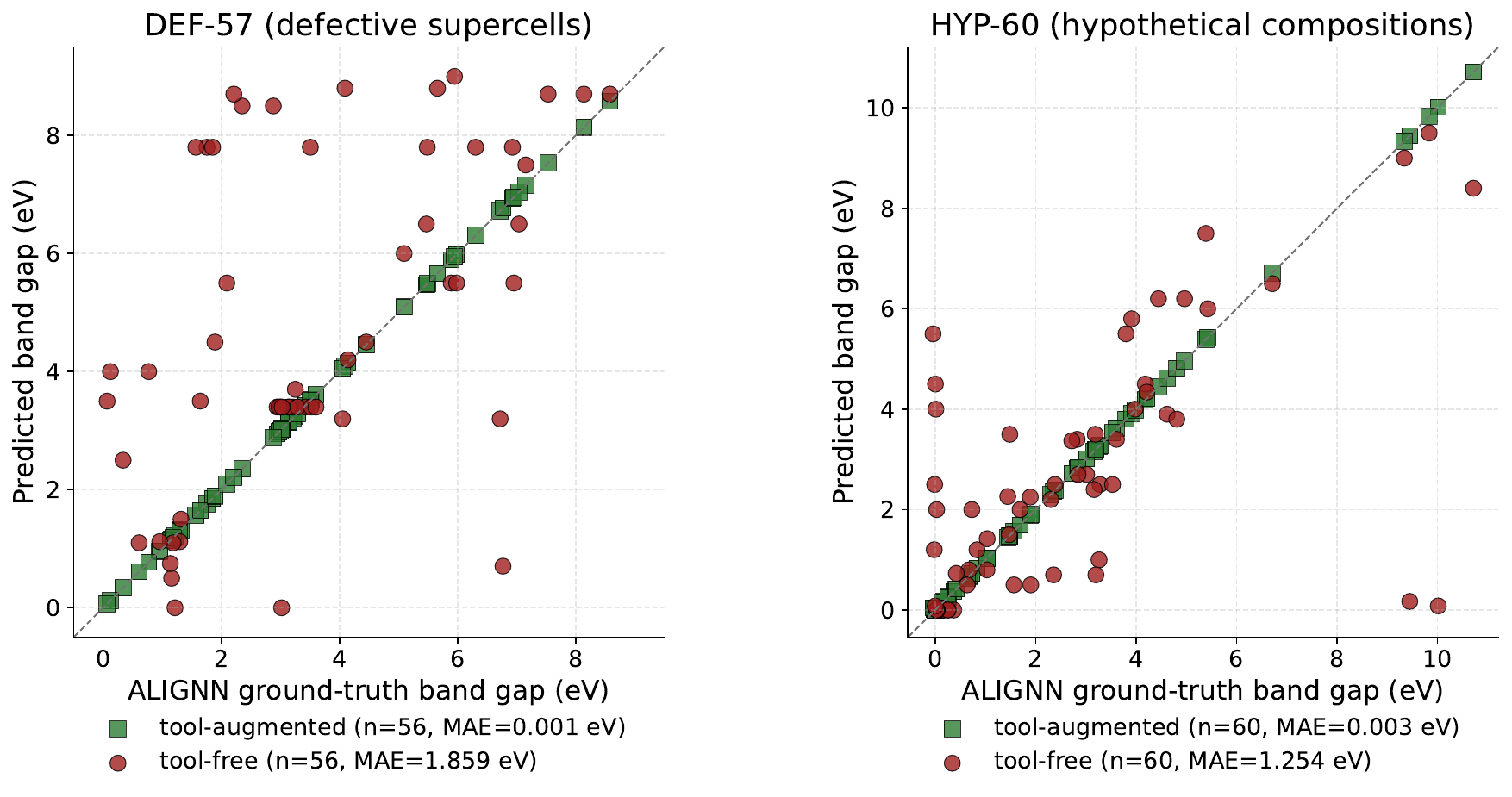}
\caption{\new{Memorization-resistant evaluation parity plots. (a) DEF-57 (57 defective supercells across five canonical hosts). (b) HYP-60 (60 hypothetical compositions from six prototype crystals). Green squares: tool-augmented agent (MAE $0.001$~eV on DEF-57, $0.003$~eV on HYP-60). Red circles: tool-free agent (MAE $1.86$~eV on DEF-57, $1.25$~eV on HYP-60). Dashed line: ideal predictions.}}
\label{fig:novel}
\end{figure}

\new{Figure~\ref{fig:novel} reports the result. On DEF-57, the tool-augmented agent reproduces the ALIGNN ground-truth bandgap to within numerical precision ($\mathrm{MAE} = 0.001$~eV) while the tool-free baseline collapses to $\mathrm{MAE} = 1.86$~eV. On HYP-60, tool-augmented $\mathrm{MAE} = 0.003$~eV versus tool-free $\mathrm{MAE} = 1.25$~eV. Visually the tool-free predictions cluster around a fixed value close to the host material's textbook bandgap, regardless of the actual defective or substituted structure, confirming the parametric-recitation failure mode. Some tasks require explicit physics-based simulation, for example full electronic band structures, relaxed atomic geometries, and X-ray diffraction patterns from arbitrary structures. These tasks therefore cannot be answered without tools regardless of LLM capability, independent of how parametric memorization shapes the canonical-materials baseline. We note that the tool-augmented MAE being on the diagonal in this evaluation is a property of the circular ground-truth choice (ALIGNN both as tool and as reference). An independent ground truth (e.g., self-consistent DFT) would lift the diagonal but would not change the load-bearing tool-free result.} Agentic systems therefore should not blindly invoke tools for every query. They should dynamically assess whether tool access is likely to improve or degrade accuracy for the specific property and material at hand.

\textbf{Toward Adaptive Tool Selection.} \new{Because the net benefit of tool augmentation is governed by database-functional faithfulness rather than by agent behavior, the decomposition of Evaluations~A--C gives a direct recipe for adaptive tool selection: the agent should invoke a database tool when the expected \textit{database-vs-experiment residual} for the target property is smaller than the expected parametric-knowledge error.} We envision a mechanism evaluating: (1) a \textit{database coverage score} that captures whether the target material and property exist in the database at the best-available level of theory, (2) a \textit{property reliability score} \new{that encodes the historical magnitude of the database-vs-experiment residual for each property--functional pair (directly measurable from the decomposition in Figure~\ref{fig:decomp})}, and (3) \textit{LLM confidence calibration} that assesses whether the LLM's parametric response is likely to be accurate based on consistency across multiple low-temperature samples. When coverage is low, \new{the residual is large}, or LLM confidence is high, the agent would bypass tools and rely on parametric knowledge. We are actively developing and benchmarking such mechanisms and anticipate that adaptive selection will outperform both always-use-tools and never-use-tools strategies.

\textbf{Autonomous Multi-Tool Workflows.} Beyond single-property prediction, the true power of AGAPI emerges in multi-tool workflows requiring coordinated execution of multiple operations. Such workflows go beyond rigid manual stitching of tools with explicit if-else statements, allowing the LLM to dynamically plan and adapt execution based on intermediate results.

Figure~S8 demonstrates a 10-operation semiconductor defect analysis pipeline, executed autonomously in response to a single natural language prompt. The workflow proceeds as follows. (1) The agent searches the JARVIS-DFT database for all GaN entries and identifies 5 polymorphs. (2) It retrieves the POSCAR structure for the most thermodynamically stable entry (JVASP-30, wurtzite $P6_3mc$, formation energy $-0.571$~eV/atom). (3) It constructs a $2 \times 1 \times 1$ supercell containing 8 atoms (4 Ga, 4 N). (4) It substitutes one Ga atom with Al to create an Al$_{0.25}$Ga$_{0.75}$N defect structure. (5) It relaxes the structure using ALIGNN-FF, confirming convergence. (6) It generates a simulated powder XRD pattern using Cu K$\alpha$ radiation ($\lambda = 1.54184$~\AA) and reports the 10 strongest peaks. (7) On successful relaxation it predicts ALIGNN-based properties for the Al-doped supercell (formation energy, MBJ and OptB88vdW bandgaps, bulk and shear moduli). (8) It computes a tight-binding band structure for the same supercell using SlaKoNet. (9) It synthesizes all results into a unified summary. (10) It generates a formatted comparison table.% NOTE: figure moved to SI (Figure S8)

\begin{figure}[H]
\centering
\includegraphics[width=0.9\textwidth]{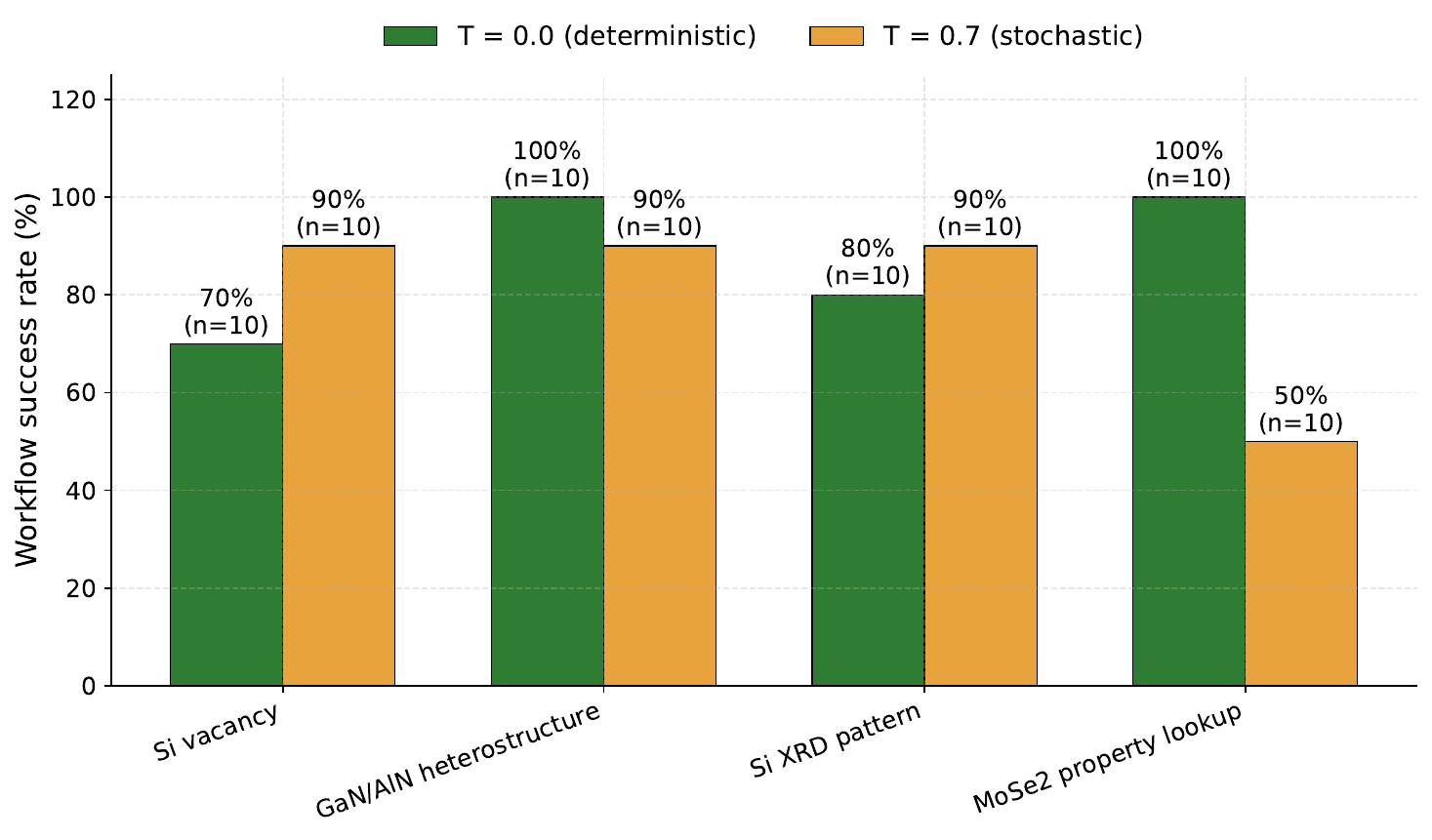}
\caption{\new{Workflow success rate per workflow under deterministic ($T = 0$, green) and stochastic ($T = 0.7$, gold) decoding. Four canonical multi-step workflows were each run $N = 10$ times at each temperature. Mean success rate is 87.5\% at $T = 0$ and 80\% at $T = 0.7$ (40 trials each).}}
\label{fig:robust}
\end{figure}

Additional demonstrated workflows include the following. (i) Autonomous GaN/AlN heterostructure interface construction, where the agent searches for both materials, identifies optimal polymorphs, and generates \new{a 32-atom (0001)/(0001) wurtzite interface (8~Al, 8~Ga, 16~N) using coincidence site lattice matching. The compact cell reflects the high-symmetry lattice match between wurtzite GaN and AlN and is reproduced in $10$ of $10$ deterministic trials in the Workflow Robustness benchmark (Figure~\ref{fig:robust})}. (ii) Powder XRD pattern generation and analysis for wurtzite GaN, where the agent retrieves the structure, simulates the diffraction pattern, identifies the 8 strongest peaks with their 2$\theta$ positions, relative intensities, and $d$-spacings, and provides crystallographic interpretation identifying dominant reflections and their Miller indices. Workflow-trace transcripts for these demonstrations are provided in the Supporting Information (Figures S1--S4).

\new{\textbf{Workflow Robustness Under Stochastic Decoding.} The demonstrations above show what AGAPI can do on curated prompts. They do not establish how often it does it across repeated runs. To quantify robustness we selected four canonical multi-step workflows (Si vacancy + bandgap prediction, GaN/AlN heterostructure interface generation, Si XRD pattern computation, and MoSe$_2$ property lookup), anchored in the integration tests of the open-source release and in the canonical workflow templates of the system prompt, and ran each $N = 10$ times at two sampling temperatures: $T = 0$ (deterministic baseline) and $T = 0.7$ (realistic stochastic-decoding stress test matching typical commercial LLM defaults). Each trial is judged a success if the agent's final response contains the expected numeric output (bandgap with eV units, XRD peaks with 2$\theta$ positions, or an interface description) and references the requested chemistry.}

\new{Figure~\ref{fig:robust} reports the result. Across 40 trials at $T = 0$ the mean workflow success rate is 87.5\% (35/40). Across 40 trials at $T = 0.7$ it is 80\% (32/40), a modest 7.5-percentage-point degradation indicating that the agent is broadly robust to stochastic decoding. Per-workflow success: GaN/AlN heterostructure 100\%/90\% (T=0/T=0.7), Si XRD 80\%/90\%, Si vacancy 70\%/90\%, MoSe$_2$ lookup 100\%/50\%. One pattern worth noting is that the Si vacancy workflow actually improves under higher temperature. At $T = 0$, the deterministic path occasionally falls into a degenerate branch where the agent halts to ask the user for clarification instead of executing the workflow. Stochastic decoding at $T = 0.7$ sometimes samples around that branch and the workflow completes.}

\textbf{Database Search with Tool Calling.} Figure~S9 illustrates how natural language queries are automatically translated into structured API calls with grounded outputs. The query ``Find all Al$_2$O$_3$ materials'' is decomposed into a JARVIS-DFT database search that returns 14 polymorphs with JARVIS-IDs, space groups, formation energies, bulk moduli, and bandgaps (both MBJ and OptB88vdW) in a formatted table. The agent identifies the R$\bar{3}$c corundum phase (JVASP-32) as the thermodynamically most stable polymorph ($\Delta E_{\text{hull}} = 0$~eV) with the highest bulk modulus (241~GPa) and bandgap (7.57~eV MBJ) among the listed phases. Critically, the agent prioritizes TBmBJ bandgaps over OptB88vdW values through system-prompt guidance and provides physical interpretations alongside the data, including explanations of formation energy, bandgap functional differences, and stability metrics that help users, particularly those new to computational materials science, understand the significance of the results.

\begin{figure}[H]
\centering
\includegraphics[width=\textwidth]{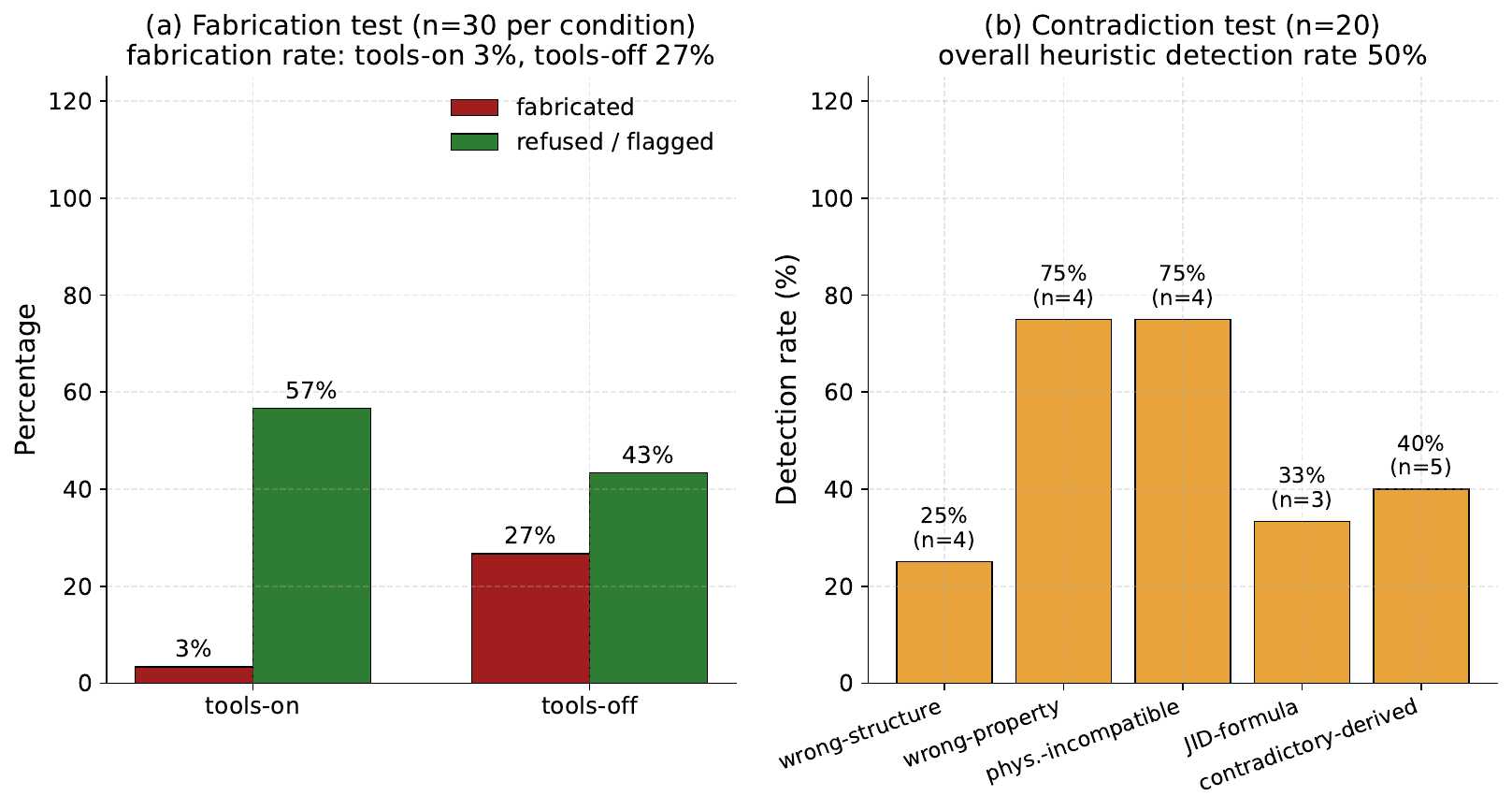}
\caption{\new{Grounding and hallucination evaluation across two complementary tests. (a) Fabrication test on 30 non-existent-material prompts (fictional JARVIS-IDs and chemically implausible compositions). With the tool layer enabled the fabrication rate is $3\%$. With it disabled the rate rises to $27\%$, a $9\times$ reduction attributable to the tool-routing design. (b) Contradiction test: per-category heuristic detection rate on the 20 deliberate-inconsistency prompts (n per category shown on each bar). Overall detection rate $50\%$ (10/20).}}
\label{fig:grounding}
\end{figure}

\new{\textbf{Grounding and Hallucination Analysis.} A central design choice of AGAPI is that the LLM is routed through a tool layer for any claim involving a database-resident value, rather than generating that value from parametric knowledge. The retrieval endpoints return explicit nulls on queries that match no record (including chemically implausible stoichiometries and fictional JARVIS-IDs), and the planner is required to surface that null rather than to fill it in from parametric memory. Evaluation~A above already establishes provenance for bulk modulus and bandgap: the agent-vs-database residual is exactly zero on every retrieved material, so $V_{\text{agent}} = V_{\text{db}}$ by construction for these two properties. To probe two complementary aspects of grounding that Evaluation~A does not cover, namely the agent's behavior on out-of-database queries and on prompts containing factual errors, we evaluate two further tests (Figure~\ref{fig:grounding}):}

\new{\textit{Fabrication test.} We constructed 30 prompts asking for properties of materials that do not exist in JARVIS-DFT (fictional JARVIS-IDs such as \texttt{JVASP-999999} and chemically implausible compositions such as \texttt{Pb$_3$As$_4$}, \texttt{FeAuPd}, \texttt{Au$_2$Pt$_3$O$_7$}) and ran each twice: once with the tool layer enabled and once with it disabled. With tools enabled the fabrication rate is $\mathbf{3\%}$ ($1/30$). Without tools it is $\mathbf{27\%}$ ($8/30$), a $9\times$ reduction in ungrounded numerical generation attributable to the tool-routing design. The remaining cases are responses we classified as ambiguous, meaning the text contained neither an extractable numeric value with units nor an explicit refusal or null statement, typically partial or hedged answers. The ambiguous rate is $40\%$ ($12/30$) tools-on and $30\%$ ($9/30$) tools-off.}

\new{\textit{Contradiction test.} We constructed 20 prompts containing a deliberate inconsistency between a user-asserted fact and the database reality, in five categories: wrong-structure claims (e.g., ``MgB$_2$ in rocksalt''), wrong-property values (``Si has bandgap of 5~eV''), physically incompatible claims (``GaAs which has Tc = 39~K''), JID-formula mismatches (``JVASP-1002 which is GaN''), and contradictory derived quantities (``Al$_2$O$_3$ corundum has dielectric constant of 2''). A keyword-based heuristic flagged $\mathbf{10/20}$ ($50\%$) of responses as surfacing the inconsistency through markers such as ``typical structure is...'', ``well-known Tc of...'', or explicit ``however''/``actually''/``in fact'' framing. Detection varied by category: wrong-property values and physically incompatible claims were flagged most reliably ($3/4$ each, $75\%$), while wrong-structure claims ($1/4$, $25\%$), JID-formula mismatches ($1/3$, $33\%$), and contradictory derived quantities ($2/5$, $40\%$) were flagged less consistently.}

\new{These measurements support a bounded claim: AGAPI substantially reduces ungrounded numerical generation, with a $9\times$ reduction in fabrication on non-existent-material prompts and nontrivial grounding-driven critical engagement with user-asserted facts.}

\textbf{Discussion.} \new{AGAPI's three-evaluation decomposition demonstrates that tool augmentation is the right design choice for agentic materials AI, with quantitatively characterizable conditions under which it most strongly outperforms tool-free LLMs. Evaluation~A shows that the agentic pipeline forwards retrieved DFT values with zero error on bulk modulus ($n = 21$) and bandgap ($n = 54$). On canonical materials whose DFT reference is well-tuned for the property of interest or for which the LLM has no strong parametric prior, tool augmentation reduces MAE against experiment by $27\%$ (bulk modulus, $7.88 \rightarrow 5.73$~GPa) and $46\%$ (dielectric constant $\varepsilon_x$, $2.88 \rightarrow 1.54$). On memorization-resistant test sets where parametric knowledge is unavailable by construction (DEF-57 and HYP-60), tool augmentation is essential: tool-augmented MAE collapses to $\leq 0.003$~eV while tool-free MAE remains at $1.25$--$1.86$~eV, with the LLM's predictions clustering around the host material's textbook bandgap regardless of the actual structure. The narrow regime in which tool augmentation can underperform tool-free LLMs is canonical materials whose memorized literature values lie closer to experiment than the retrieved DFT functional's values. Switching the retrieved DFT functional reduces the error directly (Mechanism~2: OptB88vdW $1.42$~eV $\rightarrow$ MBJ $0.48$~eV $\rightarrow$ HSE06 $0.27$~eV monotonic). Per-class stratification supports this functional-limitation picture, with per-class MAE varying from $0.13$ to $1.05$~eV and wide-bandgap classes (oxides, halides) showing the largest error. For superconducting $T_c$, the agent does not always forward the retrieved DFT value verbatim, so the tool-augmented error against experiment has two contributors: DFT functional bias and the agent-vs-database residual.}

\new{The physical intuition is that for well-characterized materials (those extensively documented in the scientific literature that forms the training corpus of modern LLMs), the LLM's parametric knowledge already encodes experimentally-close property values. In these cases, routing through a DFT database introduces an intermediary whose systematic bias against experiment can exceed the LLM's retrieval error, and the net effect against experiment is negative.} This is analogous to a domain expert consulting a computational handbook whose methods are less accurate than the experimental values the expert already knows from primary literature. \new{The situation reverses sharply for novel materials, modified structures, or computed properties (band structures, relaxed geometries) that require explicit physics-based calculation and for which the LLM has no parametric knowledge to draw upon, a regime that the DEF-57 and HYP-60 memorization-resistant evaluations above probe directly. The two regimes together (database-functional mismatch for canonical materials, demonstrated indispensability of tools for memorization-resistant ones) explain why a single aggregate benchmark can appear to favor either side depending on test-set construction.}

Several design decisions proved critical to AGAPI's utility as a research platform. First, exclusive use of self-hosted open-source LLMs eliminates cost barriers, ensures reproducibility through version pinning, and addresses intellectual property concerns inherent in commercial APIs. While commercial models may achieve marginally better performance on some benchmarks, the reproducibility and accessibility advantages of open-source models are essential for scientific applications where long-term validity matters. Second, the modular architecture separating reasoning (LLM) from execution (APIs) enables continuous improvement of both layers independently. Third, comprehensive documentation, worked examples, open-source code, and intuitive interfaces (web chatbot, Python library, voice input) reduce adoption barriers for researchers across skill levels.

Current workflows are restricted to materials for which tools exist in the API layer. \new{Within that scope, the workflow-robustness benchmark above reports a mean success rate of 87.5\% under deterministic decoding ($T = 0$) and 80\% under stochastic decoding ($T = 0.7$) across four canonical multi-step workflows and 80 trials. The 7.5-percentage-point stochastic-decoding cost is modest. On grounding and hallucination, the live Grounding and Hallucination Analysis above reports a $9\times$ reduction in fabrication on non-existent-material prompts ($3\%$ tools-on vs $27\%$ tools-off).} Comparison with related work highlights that Coscientist\cite{boiko2023autonomous} and ChemCrow\cite{bran2023chemcrow} focus primarily on organic chemistry with limited crystalline materials coverage, AtomAgents\cite{ghafarollahi2024atomagents} shares our materials focus but relies on commercial LLMs, and LLamp\cite{chiang2024llamp} provides database access but lacks physics-based property prediction tools such as band structure calculations and force-field relaxation.

The broader implications of our benchmarking results suggest that the field of agentic AI for science should move beyond capability demonstration (``can the agent do it?'') toward rigorous performance characterization (``does tool access actually improve accuracy, and under what conditions?''). This shift is essential as agentic systems are increasingly deployed in high-stakes scientific applications. Future directions include reinforcement learning for tool-selection policy optimization, multi-modal capabilities incorporating microscopy images and spectral data, collaborative multi-agent systems with specialized planning and validation agents, and expansion to additional scientific domains.

\new{In summary, we have presented a systematic evaluation of tool-augmented versus tool-free LLM predictions for materials properties, using a three-evaluation decomposition that separates the fidelity of the agentic pipeline (Evaluation~A), the choice of DFT functional underlying the retrieved entries (Evaluation~B), and the residual partitioning of experimental-reference error (Evaluation~C). For bandgap, where the agent reproduces the JARVIS-DFT entries exactly, the experimental-reference degradation is inherited DFT functional bias, not agentic malfunction. For $T_c$ the decomposition is two-component, reflecting partial parametric-knowledge substitution by the agent on this property. On the DEF-57 and HYP-60 memorization-resistant benchmarks, tool augmentation is essential: tool-free bandgap MAE rises to $1.25$--$1.86$~eV while tool-augmented MAE remains at numerical precision against the ALIGNN ground truth. Tool augmentation is therefore the load-bearing design choice for agentic materials AI: it is essential wherever parametric LLM knowledge is unavailable, and on canonical materials its benefit against experiment is governed by the choice of DFT functional underlying the retrieved entries, not by the agentic framework itself.} AGAPI provides an open-access infrastructure for these investigations, combining exclusively open-source LLMs with comprehensive materials-science tool integration and autonomous workflow orchestration. The AGAPI codebase is available at \url{https://github.com/atomgptlab/agapi}.

\section*{Data Availability}
AGAPI source code, documentation, and example workflows are available at \url{https://atomgpt.org} and \url{https://github.com/atomgptlab/agapi}.

\section*{Competing Interests}
The authors declare no competing interests.

\section*{Supporting Information}
\noindent REST API endpoint inventory (Table S1). Per-material benchmark results across the three properties reported in Evaluation~A (bandgap, bulk modulus, $T_c$) with $V_{\text{exp}}$, $V_{\text{db}}$, and $V_{\text{agent}}$ values (Table S2). Superconducting $T_c$ per-material residual decomposition (Table S3). LLM token-generation speed reference (Table S4). Software stack (Table S5). DEF-57 per-material bandgap (ALIGNN ground truth, tool-augmented, tool-free) for the 57 defective configurations (Table S6). HYP-60 per-material bandgap for the 60 hypothetical compositions (Table S7). Fabrication test: 30 non-existent-material prompts with per-condition outcomes (Table S8). Contradiction test: 20 deliberate-inconsistency prompts with category and detection (Table S9). Tool registration protocol with code excerpt (Note S1). Runnable Python API usage Jupyter notebook (Note S2, accompanying \texttt{SI\_tutorial.pdf}). Workflow-trace transcripts for the GaN/AlN heterostructure, Si vacancy bandgap, Si XRD pattern, and MoSe$_2$ property lookup demonstrations (Figures S1--S4). LLM benchmarking (Figure S5), parity plots across five properties (Figure S6), failure-mode quantitative analyses (Figure S7), 10-step Al-doped GaN autonomous workflow (Figure S8), tool-calling grounded database search (Figure S9).

\bibliography{ref}
% \clearpage
% \setcounter{figure}{0}
% \setcounter{table}{0}
% \renewcommand{\thefigure}{S\arabic{figure}}
% \renewcommand{\thetable}{S\arabic{table}}
% \input{SI.tex}
\end{document}